\documentclass[11pt]{article}

\usepackage[preprint]{acl}

\usepackage{times}
\usepackage{latexsym}

\usepackage[T1]{fontenc}

\usepackage[utf8]{inputenc}

\usepackage{microtype}

\usepackage{inconsolata}

\usepackage{graphicx}

\usepackage{algorithm}
\usepackage{algpseudocode}
\usepackage{amsmath}
\usepackage{booktabs}
\usepackage{multirow}
\usepackage[table]{xcolor}
\usepackage{makecell}
\usepackage{bbm}
\usepackage{amssymb}
\usepackage{xspace}

\usepackage{fontawesome5}
\usepackage{xcolor}

\title{Offline Exploration-Aware Fine-Tuning for\\ Long-Chain Mathematical Reasoning}

\author{\\
    {\bf Yongyu Mu\textsuperscript{1}\thanks{\xspace\xspace Work was done when Yongyu Mu was interning at Pattern Recognition Center, WeChat AI, Tencent Inc.}, Jiali Zeng\textsuperscript{2}, Fandong Meng\textsuperscript{2} \and Jingbo Zhu\textsuperscript{1}, Tong Xiao\textsuperscript{1}\thanks{\xspace\xspace Corresponding author.}} \\
    \textsuperscript{1}NLP Lab, School of Computer Science and Engineering, Northeastern University, Shenyang, China\\
    \textsuperscript{2}Pattern Recognition Center, WeChat AI, Tencent Inc, China\\
    \texttt{lixiaoyumu9@gmail.com} \\
    \texttt{\{lemonzeng,fandongmeng\}@tencent.com} \\
    \texttt{\{xiaotong,zhujingbo\}@mail.neu.edu.cn}
}

\begin{document}
\maketitle
\begin{abstract}
Through encouraging self-exploration, reinforcement learning from verifiable rewards (RLVR) has significantly advanced the mathematical reasoning capabilities of large language models. As the starting point for RLVR, the capacity of supervised fine-tuning (SFT) to memorize new chain-of-thought trajectories provides a crucial initialization that shapes the subsequent exploration landscape. However, existing research primarily focuses on facilitating exploration during RLVR training, leaving exploration-aware SFT under-explored. To bridge this gap, we propose \textbf{O}ffline e\textbf{X}ploration-\textbf{A}ware (OXA) fine-tuning. Specifically, OXA optimizes two objectives: promoting low-confidence verified teacher-distillation data to internalize previously uncaptured reasoning patterns, and suppressing high-confidence incorrect self-distillation data to redistribute probability mass of incorrect patterns toward potentially correct candidates. Experimental results across 6 benchmarks show that OXA consistently improves mathematical reasoning performance, especially achieving an average gain of $+6$ Pass@1 and $+5$ Pass@$k$ points compared to conventional SFT on the \texttt{Qwen2.5-1.5B-Math}. Crucially, OXA elevates initial policy entropy, and performance gains persist throughout extensive RLVR training, demonstrating the long-term value of OXA.
\end{abstract}

\section{Introduction}
The mathematical reasoning capabilities of large language models (LLMs) have witnessed a breakthrough by scaling up inference computation for long-chain reasoning. Building upon pre-trained LLMs, a common strategy to achieve state-of-the-art performance involves a two-stage training pipeline \cite{DBLP:journals/corr/abs-2501-12948,DBLP:journals/corr/abs-2505-09388}: (1) Supervised fine-tuning (SFT), which distills knowledge from teacher models to activate initial reasoning capabilities and learn long-chain output formats; and (2) Reinforcement learning from verifiable rewards (RLVR), which further boosts performance by encouraging self-exploration and learning from model-generated samples.

In reinforcement learning, maintaining sufficient policy entropy to prevent premature convergence and encourage exploration is fundamental \cite{williams1991function,DBLP:journals/ml/Williams92,DBLP:journals/corr/abs-2103-06257}. In the context of RLVR, thorough exploration is particularly critical, as it allows the model to discover diverse reasoning paths and unlock greater potential. To this end, existing research has attempted to manipulate policy entropy through objective-level regularizations \cite{DBLP:journals/corr/abs-2510-10959, DBLP:journals/corr/abs-2506-14758, DBLP:journals/corr/abs-2510-02249}, fine-grained update and sampling controls \cite{DBLP:journals/corr/abs-2505-18573, DBLP:journals/corr/abs-2505-22617}, and semantic-level abstractions \cite{cao2025efficient}. However, these efforts primarily focus on facilitating exploration during the RLVR process, overlooking the role of the SFT stage. As the starting point for RLVR, SFT provides a crucial initialization that shapes the subsequent exploration landscape.

Moreover, recent studies reveal that while RLVR excels at optimizing known paths, it struggles to expand the model's fundamental reasoning space. In contrast, SFT is highly effective at enabling models to internalize new reasoning pathways \cite{DBLP:journals/corr/abs-2504-13837,DBLP:journals/corr/abs-2505-14216,DBLP:conf/icml/ChuZYTXSLL025}. This suggests that by enriching the model's exploration space with diverse reasoning pathways, intuitively, SFT can facilitate exploration in the RLVR process. Despite this potential, current long-chain reasoning SFT research focuses exclusively on activating reasoning capabilities \cite{DBLP:journals/corr/abs-2506-13284,DBLP:journals/corr/abs-2506-04178} or data pruning for efficiency \cite{DBLP:journals/corr/abs-2501-19393,DBLP:journals/corr/abs-2505-17266}. This work addresses the question: \textit{How can we train exploration-engaged models for RLVR via fine-tuning?}

We envision that an ideal initialization for RLVR should exhibit two key characteristics: \textit{superior initial reasoning accuracy} and \textit{high initial policy entropy}. To achieve this, we propose \textbf{O}ffline e\textbf{X}ploration-\textbf{A}ware (OXA) fine-tuning, an algorithm designed to train on strategically selected offline reasoning trajectories. To counteract the entropy collapse illustrated in Figure \ref{fig:main_entropy}, OXA reinforces low-probability trajectories while weakening high-probability ones to preserve the smoothness of the predictive distribution. Specifically, OXA optimizes two objectives: \textit{promoting low-confidence verified teacher-distillation data} and \textit{suppressing high-confidence incorrect self-distillation data}. The former internalizes previously uncaptured reasoning trajectories, while the latter redistributes probability mass of incorrect paths toward potentially correct candidates. Since these objectives are decoupled, we introduce two variants: a base version of OXA utilizing only the first objective and the full OXA framework. While the base version accounts for the primary performance gains, the full framework synergizes superior performance with robust exploration potential.

We evaluate OXA by applying the SFT-then-RLVR paradigm to 4 LLMs ranging from 1.5B to 7B parameters. Experimental results across 6 typical mathematical benchmarks show that OXA consistently enhances reasoning capabilities. Notably, it achieves an average improvement of $+6$ Pass@1 and $+5$ Pass@$k$ points compared to conventional SFT on the 1.5B LLM. Comprehensive analysis further demonstrates that OXA not only improves performance on challenging problems but also significantly expands the reasoning output space, achieving high initial policy entropy. Crucially, these gains persist throughout extensive RLVR training and are orthogonal to existing RLVR-enhancement methods, yielding consistent additive improvements. \faGithub\ GitHub: \url{https://github.com/takagi97/OXA-Fine-tuning}

\begin{figure*}
\centering
\includegraphics[width=0.9\textwidth]{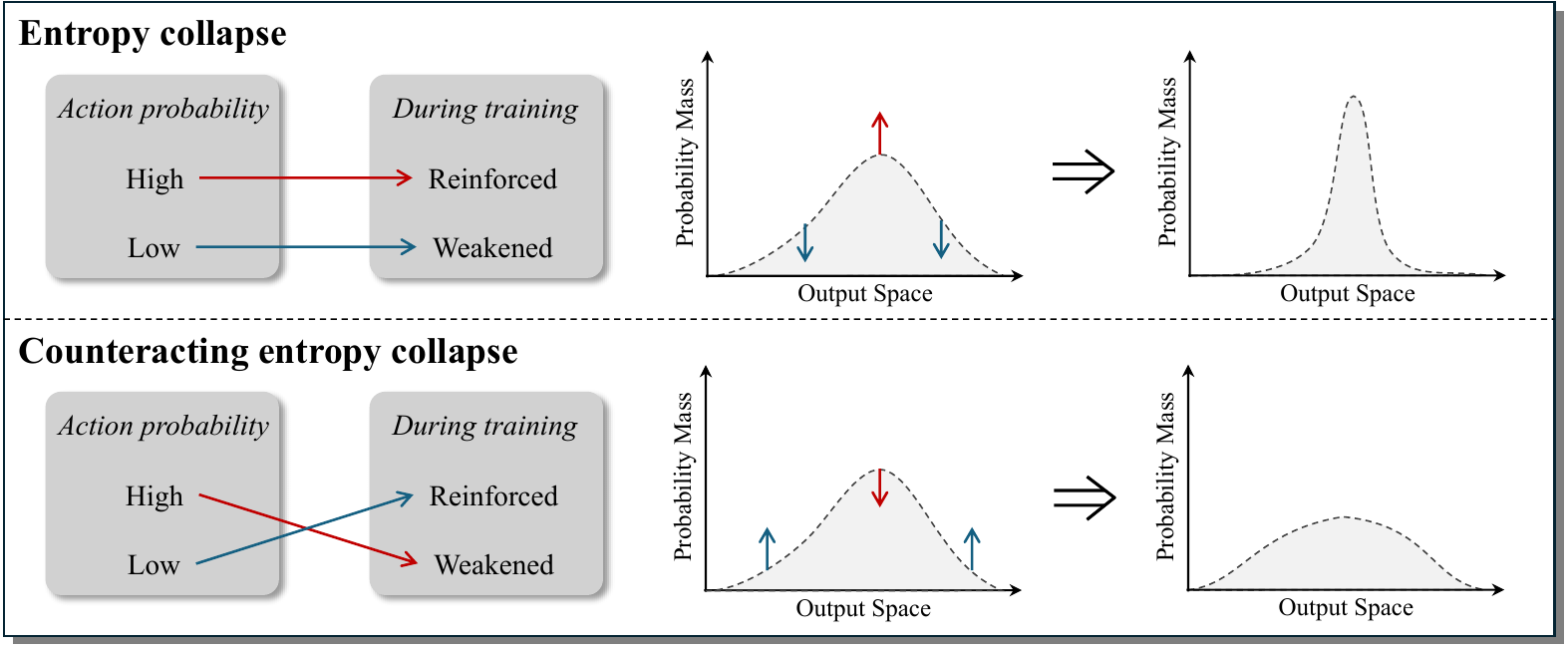}
\caption{Conceptual illustration of entropy collapse versus counteracting entropy collapse.}
\label{fig:main_entropy}
\end{figure*}

\section{Related Work}

\paragraph{Training long-chain reasoning LLMs.} There are primarily three trajectories for training long-chain reasoning LLMs. Initial efforts face the scarcity of supervised data with annotated reasoning steps. They leverage RLVR to directly train pre-trained LLMs to explore self-discovered reasoning paths toward correct answers \cite{DBLP:journals/corr/abs-2501-12948,DBLP:journals/corr/abs-2501-12599}. This base-model training process is termed Zero-RL. However, these models often suffer from poor readability and lower performance ceilings \cite{DBLP:journals/corr/abs-2501-12948}. A second paradigm involves two stages: first distilling knowledge from teacher models (e.g., Zero-RL models) into pre-trained LLMs via SFT for a cold start, then followed by RLVR. Other works extend this by integrating supervised signals from teachers into RLVR to learn superior reasoning trajectories \cite{DBLP:journals/corr/abs-2504-14945,DBLP:journals/corr/abs-2506-02208,DBLP:journals/corr/abs-2509-04419,DBLP:journals/corr/abs-2508-11408}. In this work, we follow the SFT-then-RLVR paradigm, widely validated by various open-source LLMs \cite{DBLP:journals/corr/abs-2501-12948,DBLP:journals/corr/abs-2505-09388,DBLP:journals/corr/abs-2505-07608}.

\paragraph{Analysis of SFT and RLVR.} Although both SFT and RLVR can train long-chain reasoning models, recent works reveal that their learning mechanisms differ. RLVR exhibits a generalization pattern: constrained by self-generated training samples, it improves accuracy by increasing the probability of correct answers, but fails to expand capability by sampling correct answers outside the model's output space \cite{DBLP:journals/corr/abs-2504-13837,DBLP:journals/corr/abs-2506-04913}. In contrast, SFT tends to memorize training data. By introducing new knowledge during distillation, SFT can enhance the model's capabilities on difficult problems \cite{DBLP:journals/corr/abs-2505-14216,DBLP:conf/icml/ChuZYTXSLL025}.

\paragraph{Policy entropy in RLVR.} Rooted in information theory, entropy provides a principled mechanism to manage the exploitation-exploration tradeoff. Higher policy entropy indicates that the model is more likely to explore diverse reasoning paths, thereby enhancing performance. To maintain high entropy during early training, various efforts have been made, including restricting the update of high-covariance tokens \cite{DBLP:journals/corr/abs-2505-22617}, adjusting entropy regularization coefficients \cite{DBLP:journals/corr/abs-2510-10959}, tuning rollout temperature \cite{DBLP:journals/corr/abs-2505-18573}, adding extra entropy terms to advantage calculations \cite{DBLP:journals/corr/abs-2506-14758}, leveraging cumulative entropy regulation \cite{DBLP:journals/corr/abs-2510-02249}, and elevating entropy control from the token to the semantic level \cite{cao2025efficient}. However, previous works predominantly manipulate RLVR training dynamics but overlook SFT, leaving an open question: \textit{How can we encourage models to explore more reasoning trajectories via SFT?}

\paragraph{Long-chain reasoning SFT.} Recent SFT research targeting long-chain reasoning generally follows two directions. The first focuses on eliciting reasoning capabilities in pretrained LLMs. Some studies expand reasoning paths by multi-sample distillation from teacher models \cite{DBLP:journals/corr/abs-2506-13284,DBLP:journals/corr/abs-2506-04178}, while others employ curriculum learning or decompose complex structures \cite{DBLP:journals/corr/abs-2503-10460,an-etal-2025-long,DBLP:journals/corr/abs-2503-16385}. The second direction focuses on data pruning to enhance training efficiency \cite{DBLP:journals/corr/abs-2501-19393,DBLP:journals/corr/abs-2505-17266}. While our method also involves data selection, it is distinct in its objective: rather than optimizing for training efficiency, OXA aims to cultivate exploration-engaged models that provide a superior initialization for subsequent RLVR.

\section{Methodology}
\label{sec:Methodology}
In the SFT-then-RLVR training paradigm, SFT models serve as the critical foundation for subsequent reinforcement learning. We hypothesize that an ideal SFT model should exhibit superior initial reasoning capabilities while simultaneously fostering exploration-engaged behavior to unlock greater RL potential. Specifically, we aim to achieve two objectives:
\begin{itemize}
    \item \textbf{Achieving higher initial performance:} Providing a robust backbone that maintains or amplifies the performance superiority established during SFT throughout the RLVR process.
    \item \textbf{Maintaining higher policy entropy:} Broadening the reasoning output space and facilitating sampling diverse reasoning trajectories.
\end{itemize}
To this end, we propose \textbf{O}ffline e\textbf{X}ploration-\textbf{A}ware \textbf{(OXA)} fine-tuning, an algorithm designed to train on strategically selected offline reasoning trajectories. Specifically, inspired by counteracting entropy collapse, OXA optimizes the model by promoting low-confidence correct answers and suppressing high-confidence errors. The former increases the likelihood of generating previously uncaptured reasoning trajectories, particularly those near the distribution boundaries, while the latter redistributes probability mass of incorrect reasoning paths toward other potentially correct candidates. Ultimately, OXA yields a model that combines enhanced performance with high initial policy entropy.

As a purely SFT-based approach, OXA offers two distinct advantages: First, it enhances the exploration capability without changing the RLVR framework, thereby preserving training stability while delivering consistent performance gains. Second, empirical results demonstrate that OXA is orthogonal to RLVR-enhancement methods, providing additive improvements.

\subsection{Dissecting Entropy Dynamics}
Policy entropy, quantifying the smoothness of the predictive distribution of model $\pi_{\theta}$, is defined as:
\begin{align}
\mathcal{H}(\pi_{\theta}) = - \sum_{i=1}^{|\mathcal{V}|} p_i \log p_i,
\end{align}
where $p_i$ represents the probability of the $i$-th token in the vocabulary $\mathcal{V}$ of size $|\mathcal{V}|$. In the context of training long-chain mathematical reasoning LLMs, entropy serves as a direct proxy for the diversity of reasoning paths the model can sample. A higher policy entropy indicates a more uniform distribution of probability mass, enabling the model to explore more candidate outputs. Conversely, a lower entropy characterizes a sharp distribution where the model's confidence is excessively concentrated on a limited set of tokens, thereby restricting its exploration space and limiting the variety of generated reasoning trajectories.

\paragraph{Entropy collapse.} When a model is trained to convergence, the entropy generally decreases. As illustrated in the upper one of Figure \ref{fig:main_entropy}, this phenomenon stems from the model being highly aligned with the empirical distribution of the training data, including reinforcing the distribution peaks while suppressing the troughs.

\paragraph{Counteracting entropy collapse.} To mitigate entropy collapse, a straightforward strategy is to inversely influence the distribution dynamics: promoting the probability mass at the distribution troughs while suppressing over-confident peaks, which is depicted in the lower one of Figure \ref{fig:main_entropy}.

\subsection{Offline Exploration-Aware Fine-tuning}
A higher policy entropy is preferred for exploration-engaged models. However, directly reinforcing the probability at the trough and weakening that at the peak can severely destabilize the model. OXA provides a solution that selectively promotes low-confidence truths and suppresses high-confidence errors. All reasoning instruction data for this process is curated offline.

\subsubsection{Promote Low-Confidence Truths}
This objective aims to reinforce the model's probability mass in low-confidence regions by training on verified teacher-distilled data via the maximum likelihood estimation (MLE) criterion. Through this process, the model internalizes previously uncaptured reasoning trajectories, thereby effectively expanding its reasoning space.

\paragraph{Data.} Based on the teacher-distilled dataset, we first employ the rule-based verifier used during RLVR training to filter the data, retaining only correct reasoning paths. We then use perplexity (PPL) to quantify the model's confidence in a specific reasoning route. For a reasoning trajectory $S = \{s_1, s_2, \dots, s_K\}$, the PPL is defined as:
\begin{equation}
\text{PPL}(S) = \exp \left( -\frac{1}{K} \sum_{t=1}^{K} \log p(s_t \mid s_{<t}) \right),
\end{equation}
where $K$ denotes the sequence length and $p(s_t \mid s_{<t})$ is the conditional probability assigned by the model $\pi_{\theta}$. A higher PPL indicates that the model is less confident in the trajectory, signifying a hard-to-sample reasoning path, while a lower PPL suggests the opposite.

\begin{algorithm}[t]
\small
\caption{Gaussian-Guided PPL Sampling}
\label{alg:gaussian_sampling}
\textbf{Input:} Dataset $\mathcal{D} = \{(q, r, p, l)\}$ containing query, response, PPL, and length; Hyperparameters $\mu, \sigma$, total size $N$, maximum responses per query $d$, PPL range $[p_{\min}, p_{\max}]$ and bin width $w$. \\
\textbf{Output:} Selected subset $\mathcal{R}$.

\begin{algorithmic}[1]
\State \textbf{Step 1: Binning}
\State Define $M = \lceil (p_{\max} - p_{\min}) / w \rceil$ bins.
\State Partition $\mathcal{D}$ into bins $\mathcal{B}_1, \dots, \mathcal{B}_M$ based on PPL $p$.
\State Discard samples where $p \notin [p_{\min}, p_{\max}]$.

\State \textbf{Step 2: Target Distribution Setup}
\For{each bin $i \in \{1, \dots, M\}$}
    \State Let $x_i$ be the center PPL of bin $i$.
    \State Compute density $d_i = \frac{1}{\sigma\sqrt{2\pi}} e^{-\frac{1}{2}(\frac{x_i - \mu}{\sigma})^2}$.
\EndFor
\State Normalize counts: $T_i \leftarrow \lfloor N \cdot (d_i / \sum_{j} d_j) \rfloor$.

\State \textbf{Step 3: Length-Prioritized Sampling}
\For{each bin $i \in \{1, \dots, M\}$}
    \State Sort $\mathcal{B}_i$ by length $l$ in descending order.
    \State Initialize bin counter $c_i \leftarrow 0$.
    \For{each candidate $(q, r, p, l) \in \mathcal{B}_i$}
        \If{$c_i < T_i$ \textbf{and} $Count(q) < d$}
            \State $\mathcal{S} \leftarrow \mathcal{S} \cup \{(q, r)\}$.
            \State $Count(q) \leftarrow Count(q) + 1$, $c_i \leftarrow c_i + 1$.
        \EndIf
    \EndFor
\EndFor
\State \Return $\mathcal{R}$
\end{algorithmic}
\end{algorithm}

To prevent the training set from being dominated by excessively difficult learning samples, we design a Gaussian-guided PPL sampling algorithm (Algorithm \ref{alg:gaussian_sampling}). This algorithm samples data according to a predefined Gaussian PPL distribution consisting of three key stages: binning, target distribution setup, and length-prioritized sampling. It allows us to explicitly control the PPL distribution centered at $\mu$ with a dispersion $\sigma$, while enforcing maximum responses per query $d$. Specifically, an increase in $\mu$ shifts the selection toward higher-PPL reasoning paths, while $\sigma$ modulates the sampling density for data points deviating from the central perplexity. Furthermore, within the same PPL bin, we prioritize longer responses to enhance the model's capability in generating complex, multi-step reasoning chains. Hyperparameter details for this sampling process are provided in Appendix \ref{app:Preliminary Experiments}.

\paragraph{Training.} Given a batch of $M$ reasoning trajectories $\mathcal{B}_{\text{MLE}} = \{S_1, S_2, \dots, S_{M}\}$ (each with $K_S$ tokens) selected for promotion, we adopt the MLE objective by minimizing the cross-entropy loss:
\begin{align}
\mathcal{L}_{\text{CE}} = - \frac{1}{M \cdot K_S} \sum_{S \in \mathcal{B}_{\text{MLE}}} \sum_{t=1}^{K_S} \log p(s_t \mid s_{<t}).
\end{align}

\subsubsection{Suppress High-Confidence Errors}
The second objective of OXA focuses on weakening the probability mass at erroneous peaks by suppressing high-confidence but incorrect reasoning trajectories via the unlikelihood loss. This process redistributes the probability mass from incorrect paths toward potentially correct alternatives.

\paragraph{Data.} Since the rollouts from a pre-trained LLM often suffer from low quality, we first train an instruction-following model using a small set of teacher-distillation data, then use it to sample reasoning trajectories. After verifying trajectories, we calculate the PPL of incorrect ones using the pre-trained LLM to assess confidence. Subsequently, we select the samples with the lowest PPL that fail the verification for suppression.

\paragraph{Training.} Given a batch of $N$ reasoning trajectories $\mathcal{B}_{\text{UL}} = \{S_1, S_2, \dots, S_{N}\}$ identified as high-confidence errors, we apply the token-level unlikelihood loss \cite{DBLP:conf/iclr/WelleckKRDCW20}:
\begin{align}
\mathcal{L}_{\text{UL}} = - \frac{1}{N \cdot K_S} \sum_{S \in \mathcal{B}_{\text{UL}}} \sum_{t=1}^{K_S} \log \left(1 - p(s_t \mid s_{<t})\right).
\end{align}

\subsubsection{Global Training Objective}
Combining these components, the final OXA objective integrates both losses. We introduce a hyperparameter $\alpha$ to weight the unlikelihood loss:
\begin{align}
\mathcal{L} = \mathcal{L}_{\text{CE}} + \alpha \cdot \mathcal{L}_{\text{UL}}.
\end{align}
In practice, $\alpha$ is kept small to prevent excessive gradient magnitudes that could destabilize training. A theoretical analysis demonstrating how a large $\alpha$ leads to vanishing or exploding gradients is provided in Appendix \ref{app:Theoretical Analysis of Unlikelihood Loss}. 

We evaluate two variants of our framework: a baseline version using only the first objective (OXA$_{\text{MLE}}$) and the complete framework (OXA$_{\text{Full}}$).

\begin{table*}[ht]
\LARGE
\begin{center}
\centering
\resizebox{1.0\textwidth}{!}{
\begin{tabular}{l|cc|cc|cc|cc|cc|cc|cc}
\toprule
\multirow{2.5}{*}{\textbf{Model}} & \multicolumn{2}{c|}{\textbf{AIME24}} & \multicolumn{2}{c|}{\textbf{AIME25}} & \multicolumn{2}{c|}{\textbf{BRUMO25}} & \multicolumn{2}{c|}{\textbf{CMIMC25}} & \multicolumn{2}{c|}{\textbf{HMMT25}} & \multicolumn{2}{c|}{\textbf{Minerva}} & \multicolumn{2}{c}{\textbf{Avg. Perf.}} \\
\cmidrule(lr){2-3} \cmidrule(lr){4-5} \cmidrule(lr){6-7} \cmidrule(lr){8-9} \cmidrule(lr){10-11} \cmidrule(lr){12-13} \cmidrule(lr){14-15}
 & Pass@1 & Pass@128 & Pass@1 & Pass@128 & Pass@1 & Pass@128 & Pass@1 & Pass@128 & Pass@1 & Pass@128 & Pass@1 & Pass@64 & Pass@1 & Pass@$k$ \\
\midrule
\multicolumn{15}{c}{\textit{Qwen2.5-1.5B-Math}} \\
\midrule
Base & 9.4 & 53.3 & 4.5 & 36.7 & 11.5 & 40.0 & 2.4 & 35.0 & 0.5 & 26.7 & 11.9 & 62.9 & 6.7 & 42.4 \\
\midrule
SFT$_{\text{LP}}$ & 20.5 & 76.7 & 20.5 & 53.3 & 29.8 & 73.3 & 9.8 & 50.0 & 7.1 & 40.0 & 20.7 & 61.4 & 18.1 & 59.1 \\
SFT & 23.2 & 80.0 & 23.8 & 60.0 & 29.0 & 80.0 & 11.0 & 52.5 & 11.6 & 50.0 & 22.3 & 61.4 & 20.2 & 64.0 \\
\rowcolor{gray!20} OXA$_{\text{MLE}}$ & 35.0 & \textbf{83.3} & \textbf{27.1} & \textbf{66.7} & \textbf{36.1} & \textbf{83.3} & 14.9 & 57.5 & \textbf{18.5} & \textbf{60.0} & \textbf{25.7} & 63.2 & \textbf{26.2} & 69.0 \\
\rowcolor{gray!20} OXA$_{\text{Full}}$ & \textbf{35.4} & 80.0 & 26.7 & \textbf{66.7} & 34.7 & \textbf{83.3} & \textbf{15.9} & \textbf{62.5} & 18.2 & \textbf{60.0} & 22.8 & \textbf{64.0} & 25.6 & \textbf{69.4} \\
\midrule
SFT$_{\text{LP}}$\dag & 22.3 & 70.0 & 22.6 & 60.0 & 29.4 & 76.7 & 10.7 & 45.0 & 9.5 & 46.7 & 22.0 & 58.5 & 19.4 & 59.5 \\
SFT\dag & 27.2 & 76.7 & 25.4 & 60.0 & 34.6 & 76.7 & 11.9 & 45.0 & 13.8 & 56.7 & 23.3 & 62.5 & 22.7 & 62.9 \\
\rowcolor{gray!20} OXA$_{\text{MLE}}$\dag & \textbf{40.1} & \textbf{83.3} & 27.9 & \textbf{63.3} & 39.1 & \textbf{83.3} & 16.4 & \textbf{60.0} & \textbf{19.7} & 53.3 & 27.1 & \textbf{65.8} & 28.4 & 68.2 \\
\rowcolor{gray!20} OXA$_{\text{Full}}$\dag & 39.0 & \textbf{83.3} & \textbf{29.3} & \textbf{63.3} & \textbf{41.8} & 76.7 & \textbf{17.4} & 57.5 & 19.1 & \textbf{66.7} & \textbf{27.6} & 65.1 & \textbf{29.0} & \textbf{68.8} \\

\midrule
\multicolumn{15}{c}{\textit{Qwen2.5-7B-Math}} \\
\midrule
Base & 16.0 & 63.3 & 7.3 & 43.3 & 9.8 & 53.3 & 2.4 & 35.0 & 0.6 & 16.7 & 13.2 & 64.0 & 8.2 & 45.9 \\
\midrule
SFT$_{\text{LP}}$ & 38.4 & 83.3 & 29.0 & 63.3 & 41.2 & 86.7 & 20.5 & 60.0 & 19.3 & 63.3 & 34.2 & 62.1 & 30.4 & 69.8 \\
SFT & 47.5 & 86.7 & 34.2 & 80.0 & 48.0 & 83.3 & 26.8 & 77.5 & 24.4 & 66.7 & 33.3 & 65.4 & 35.7 & 76.6 \\
\rowcolor{gray!20} OXA$_{\text{MLE}}$ & \textbf{54.5} & \textbf{90.0} & \textbf{39.3} & \textbf{83.3} & \textbf{50.9} & \textbf{90.0} & \textbf{26.6} & 75.0 & \textbf{24.5} & \textbf{80.0} & \textbf{37.3} & \textbf{65.4} & \textbf{38.8} & \textbf{80.6} \\
\rowcolor{gray!20} OXA$_{\text{Full}}$ & 50.2 & 86.7 & 36.7 & \textbf{83.3} & \textbf{50.9} & \textbf{90.0} & 24.8 & \textbf{80.0} & 23.1 & 70.0 & 36.3 & 64.7 & 37.0 & 79.1 \\
\midrule
SFT$_{\text{LP}}$\dag & 42.0 & 76.7 & 31.6 & 66.7 & 46.1 & 80.0 & 23.3 & 55.0 & 21.3 & 56.7 & 36.7 & 63.2 & 33.5 & 66.4 \\
SFT\dag & 50.9 & \textbf{90.0} & 35.1 & 80.0 & 51.6 & 83.3 & 30.3 & \textbf{67.5} & 23.6 & 70.0 & 35.1 & 64.0 & 37.8 & 75.8 \\
\rowcolor{gray!20} OXA$_{\text{MLE}}$\dag & 57.9 & 83.3 & \textbf{42.0} & \textbf{93.3} & 54.3 & \textbf{90.0} & 28.3 & 65.0 & 26.5 & 63.3 & 38.7 & 64.3 & \textbf{41.3} & \textbf{76.6} \\
\rowcolor{gray!20} OXA$_{\text{Full}}$\dag & \textbf{58.9} & 83.3 & 40.8 & 80.0 & \textbf{54.4} & \textbf{90.0} & \textbf{28.4} & 60.0 & \textbf{26.6} & \textbf{66.7} & \textbf{38.9} & \textbf{66.9} & \textbf{41.3} & 74.5 \\

\bottomrule
\end{tabular}
}
\caption{Performance comparison of fine-tuning methods and their corresponding RLVR stages. ``Base'' denotes the pre-trained LLMs, and \dag indicates models after RLVR training. Pass@1 scores are averaged over 128 samples, except for Minerva which is averaged over 64 samples. Best results within each SFT/RLVR group are bolded.}
\label{tab:main_experiment}
\vspace{-0.9em}
\end{center}
\end{table*}

\section{Experiments}
In this section, we demonstrate that OXA is capable of evolving pre-trained LLMs into exploration-engaged RLVR starting points, possessing superior mathematical reasoning capabilities to unlock further RL potential.
\begin{itemize}
    \item In Section \ref{sec:Main_Results}, our main experiment shows that OXA fulfills two objectives: improving initial performance and maintaining higher policy entropy.
    \item In Section \ref{sec:Scaling_Analysis}, we validate the effectiveness of OXA when generalizing to two additional LLMs and scaling the training data.
    \item In Section \ref{sec:Ablation_Study}, we present extensive ablation studies, including validating each component and hyperparameter, analyzing the impact of long data, assessing generalization to out-of-distribution reasoning tasks, and demonstrating orthogonality to other methods.
\end{itemize}

\subsection{Experiment Setups}

\paragraph{Model.} We select several pre-trained LLMs as the training start points. Our main experiment is based on \texttt{Qwen2.5-1.5B-Math} and \texttt{Qwen2.5-7B-Math}, hereafter referred to as the 1.5B and 7B models, respectively. Our main experiment follows the full SFT-then-RLVR paradigm by applying different SFT methods while maintaining the unified RLVR settings. In our model generalization experiment, we also validate OXA on \texttt{LLaMA3.2-3B-Base} and \texttt{Qwen3-1.7B-Base}.

\paragraph{Baselines.} We choose two baselines. The first is conventional SFT, which uses teacher distillation data as supervised signals to fine-tune pre-trained LLMs. The second is low-PPL preferred SFT (SFT$_{\text{LP}}$), which fine-tunes models using distillation data with the lowest PPL, serving as a contrast to our approach. To ensure fairness, as OXA utilizes $50,000$ teacher distillation samples, we select an equal amount of data randomly for conventional SFT and based on low-PPL preference for SFT$_{\text{LP}}$. Furthermore, in Section \ref{sec:Ablation_Study}, we compare OXA with conventional SFT using the full dataset with 2.6 million samples.

\begin{figure*}
\centering
\includegraphics[width=1.0\textwidth]{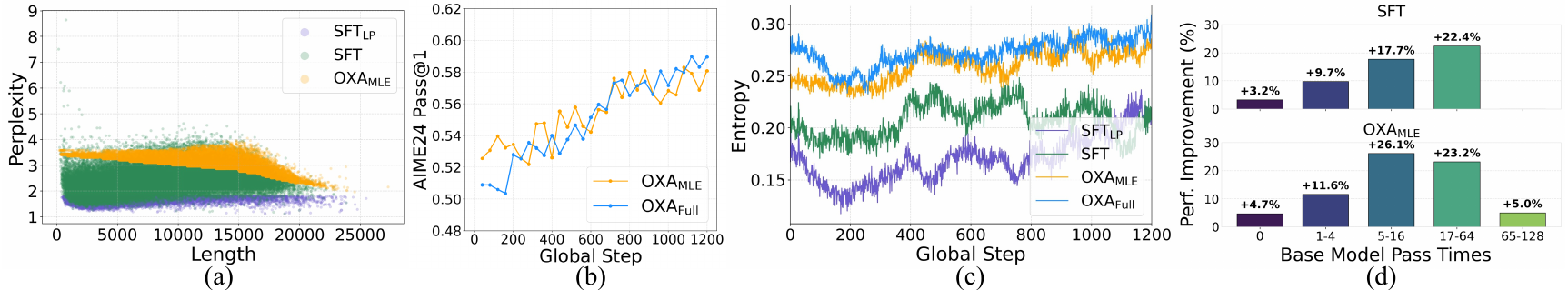}
\caption{(a) Sequence length and PPL distributions of the 7B model's training data. (b) Performance trajectories of the 7B model fine-tuned with OXA$_{\text{MLE}}$ and OXA$_{\text{Full}}$ on AIME24. (c) Policy entropy dynamics during RLVR training for various fine-tuning methods based on the 7B model. (d) Performance gains of SFT and OXA$_{\text{MLE}}$ on the 1.5B model and the Minerva benchmark, grouped by the base model's pass counts.}
\label{fig:mix_fig1}
\end{figure*}

\paragraph{SFT Details.} We use \texttt{AceReason-1.1-SFT}\footnote{\url{https://huggingface.co/datasets/nvidia/AceReason-1.1-SFT}}, a pollution-free dataset containing 2.6 million unverified DeepSeek-R1 distilled mathematical samples. After tracing back original answers, we use \texttt{math-verify}\footnote{\url{https://github.com/huggingface/Math-Verify}} to filter out incorrect reasoning paths, leaving nearly 2 million samples. We then apply our sampling algorithm to select $50,000$ high-PPL correct samples and $50,000$ low-PPL incorrect ones. We separately evaluate OXA$_{\text{MLE}}$ and OXA$_{\text{Full}}$. All of the instruction datasets maintain a one-query-to-one-response ratio. Figure \ref{fig:mix_fig1} (a) illustrates the distribution of sequence lengths and PPL for the 7B model's training data. During fine-tuning, we use a batch size of $128$ and a UL loss weight of $10^{-4}$ for $6$ epochs, with learning rates of $2.5 \times 10^{-4}$ (1.5B) and $5 \times 10^{-5}$ (7B). See Appendix \ref{app:Detailed Experimental Setup} for more details.

\paragraph{RLVR Details.} Our training dataset is a subset of DeepScaleR-40K\footnote{\url{https://huggingface.co/datasets/agentica-org/DeepScaleR-Preview-Dataset}}. For RL training, we use a maximum output length of $16,384$, $8$ rollouts per prompt, a batch size of $64$, and a decoding temperature of $0.85$. The learning rate is set to $2 \times 10^{-6}$. We train the 1.5B and 7B models for $1,600$ and $1,200$ update steps, respectively. For each RL experiment, we report results from the checkpoint achieving the peak score on the AIME24 benchmark. See Appendix \ref{app:Detailed Experimental Setup} for other details.

\paragraph{Evaluation.} We comprehensively evaluate on six mathematical benchmarks: AIME24, AIME25, BRUMO25 \cite{DBLP:journals/corr/abs-2505-23281}, CMIMC25 \cite{DBLP:journals/corr/abs-2505-23281}, HMMT25 \cite{DBLP:journals/corr/abs-2505-23281}, and Minerva \cite{DBLP:conf/nips/LewkowyczADDMRS22}. We report Pass@1 and Pass@$k$ scores to assess reasoning capabilities, where the latter represents the model's potential to solve problems. To ensure stability, Pass@1 is averaged over $k$ samples. We set $k=64$ for Minerva and $k=128$ for all other datasets. By default, we generate from the models using a temperature of $0.6$, a Top-p value of $0.95$, and a maximum output length of $32,768$ tokens.

\begin{figure*}
\centering
\includegraphics[width=1.0\textwidth]{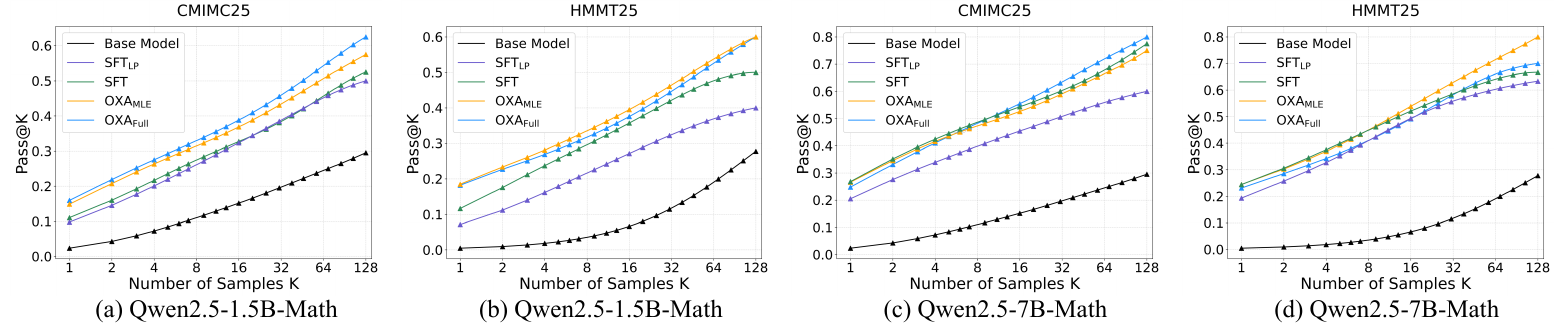}
\caption{Pass@$k$ curves of 1.5B and 7B models across different fine-tuning methods on CMIMC25 and HMMT25.}
\label{fig:passk_fig2}
\vspace{-0.9em}
\end{figure*}

\subsection{Main Results}
\label{sec:Main_Results}

\paragraph{Results of fine-tuning.} We evaluate 1.5B and 7B LLMs across four configurations: standard SFT, low-PPL SFT (the converse of OXA), and OXA in its MLE-only variant and its complete form (MLE + UL Objective), denoted as SFT, SFT$_{\text{LP}}$, OXA$_{\text{MLE}}$, and OXA$_{\text{Full}}$, respectively. As summarized in Tables \ref{tab:main_experiment} and \ref{tab:len}, several key observations emerge: (1) Both OXA variants significantly outperform the baselines; notably, the OXA fine-tuned 1.5B model achieves average improvements of $+6$ Pass@1 points and $+5$ Pass@$k$ points over conventional SFT. These gains suggest that OXA not only enhances base reasoning performance but also augments the model's capability to solve challenging problems. (2) SFT$_{\text{LP}}$ consistently underperforms relative to SFT and lags significantly behind OXA$_{\text{MLE}}$, highlighting the effectiveness of internalization low-confidence samples. (3) While OXA$_{\text{Full}}$ yields a slightly lower Pass@1 score than OXA$_{\text{MLE}}$, its Pass@$k$ performance remains competitive, indicating robust solution diversity. (4) Despite a selection bias toward longer reasoning trajectories, OXA does not substantially alter the average response length of the model.

\begin{table}[t!]
  \LARGE
  \centering
  \resizebox{0.35\textwidth}{!}{
  \begin{tabular}{l|cccc}
    \toprule
    \textbf{Model} & \textbf{SFT$_{\text{LP}}$} & \textbf{SFT} & \textbf{OXA$_{\text{MLE}}$} & \textbf{OXA$_{\text{Full}}$} \\
    \midrule
    1.5B & 15205 & 15054 & 15809 & 15700 \\
    7B   & 13477 & 12388 & 12677 & 13002 \\
    \bottomrule
  \end{tabular}
  }
  \caption{Average output lengths of fine-tuned Qwen2.5-1.5B/7B on the AIME24 benchmark.}
  \label{tab:len}
\end{table}

\paragraph{Results of reinforcement learning.} Subsequently, we perform extensive RLVR training on the fine-tuned LLMs, spanning $1,600$ update steps for the 1.5B models and $1,200$ steps for the 7B models. Based on the results marked with \dag in Table \ref{tab:main_experiment}, we observe that: (1) The performance gains achieved by OXA persist throughout the RLVR process, yielding more robust reasoning models upon completion of the full SFT-then-RLVR pipeline. (2) Compared to OXA$_{\text{MLE}}$, OXA$_{\text{Full}}$ achieves superior or competitive performance after RLVR training. Specifically, Figure \ref{fig:mix_fig1} (b) illustrates that OXA$_{\text{Full}}$ exhibits a more rapid performance ascent during the RL process, ultimately leading to higher scores.

\paragraph{Dynamics of policy entropy.} Figure \ref{fig:mix_fig1} (c) records the policy entropy dynamics for the 7B model during RLVR training. The results indicate that, compared to SFT$_{\text{LP}}$ and conventional SFT, OXA models—particularly OXA$_{\text{Full}}$—sustain higher entropy levels during the initial training phase. This validates the effectiveness of our approach in facilitating the sampling of diverse reasoning trajectories.

\begin{figure*}
\centering
\includegraphics[width=1.0\textwidth]{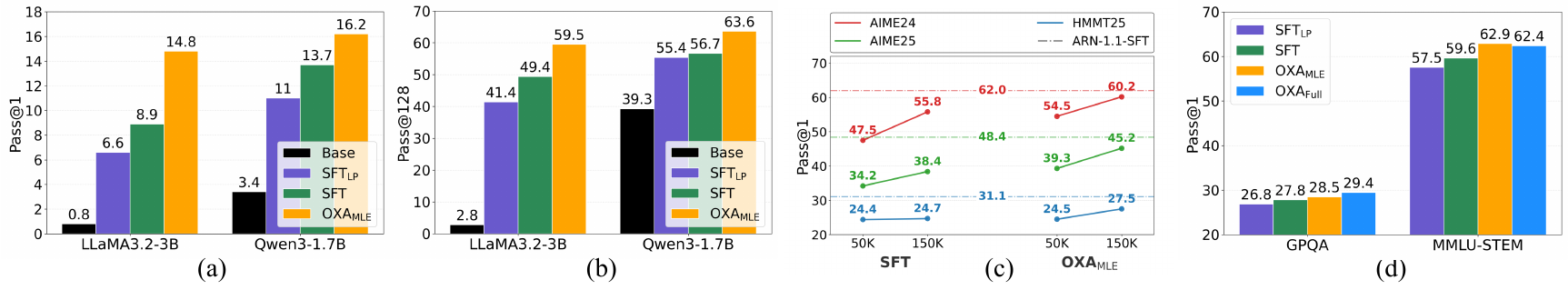}
\caption{(a)-(b) Average performance across 6 mathematical benchmarks for \texttt{LLaMA3.2-3B} and \texttt{Qwen3-1.7B} under various fine-tuning methods. (c) Scalability of SFT and $\text{OXA}_{\text{MLE}}$ as training data increases, compared against ARN-1.1-SFT—a model fine-tuned on millions of samples using the same backbone. (d) Generalization performance of 1.5B variants on out-of-domain benchmarks, including GPQA and MMLU-STEM.}
\label{fig:mix_fig2}
\vspace{-0.9em}
\end{figure*}

\paragraph{Frontiers of reasoning potential.} To understand the distinct impact of OXA compared to other SFT methods, we categorize Minerva's problems by difficulty based on the pre-trained LLM's pass counts across 128 rollouts. Figure \ref{fig:mix_fig1} (d) shows that OXA outperforms SFT across all difficulty levels, particularly in the range of 5 to 16, where more challenging problems reside, yielding significant gains. Moreover, Pass@$k$ results in Figure \ref{fig:passk_fig2} validate that OXA expands the model's reasoning potential. These results consistently demonstrate that OXA increases the likelihood of generating previously uncaptured low-probability reasoning paths, thus enabling the model to solve harder problems.

\subsection{Scaling Analysis}
\label{sec:Scaling_Analysis}

\paragraph{Model Generalization.} Beyond the Qwen2.5 series, we further evaluate the efficacy of OXA on \texttt{LLaMA3.2-3B} and \texttt{Qwen3-1.7B} models. Notably, LLaMA3.2 serves as a representative of models that have not undergone extensive pre-training on mathematics corpora. Figure \ref{fig:mix_fig2} (a) and (b) illustrate the average performance across the 6 mathematical benchmarks. Our results demonstrate that OXA consistently achieves the best results with substantial performance margins. This is particularly evident in the LLaMA3.2 model, where OXA outperforms vanilla SFT by nearly $+6$ Pass@1 points and $+10$ Pass@$k$ points. These findings suggest that OXA yields more pronounced improvements for models lacking mathematics pre-training.

\paragraph{Data Scaling.} We further investigate the scalability of OXA by increasing the training data size from $50,000$ to $150,000$ samples. We report the performance of SFT, $\text{OXA}_{\text{MLE}}$, and \texttt{ARN-1.1-SFT}—a strong baseline fine-tuned on the full \texttt{AceReason-1.1-SFT} dataset. \texttt{ARN-1.1-SFT} is trained by fine-tuning \texttt{Qwen2.5-7B-Math} with 2.6 million mathematical and 1.3 million code reasoning trajectories \cite{DBLP:journals/corr/abs-2506-13284}. As illustrated in Figure \ref{fig:mix_fig2} (c), scaling the training data significantly enhances the performance of OXA, which consistently maintains a substantial lead over the SFT baseline. Notably, with only $150,000$ samples, OXA achieves performance nearly on par with the \texttt{ARN-1.1-SFT} model, despite the latter being trained on millions of trajectories.

\begin{table}[t!]
\LARGE
    \centering
    \resizebox{1.0\linewidth}{!}{
    \begin{tabular}{llcc}
        \toprule
        \multicolumn{2}{c|}{\textbf{Configuration}} & \textbf{AIME24} & \textbf{AIME25} \\
        \midrule
        \multirow{3}{*}{\makecell[l]{Learning rate (1.5B)}} 
        & \multicolumn{1}{l|}{3.0e-4} & \textbf{23.8} & 23.1 \\
        & \multicolumn{1}{l|}{\textbf{2.5e-4}} & 23.2 & \textbf{23.8} \\
        & \multicolumn{1}{l|}{1.0e-4} & 22.8 & 23.1 \\
        \midrule
        \multirow{3}{*}{\makecell[l]{Learning rate (7B)}} 
        & \multicolumn{1}{l|}{1.0e-4} & 45.9 & \textbf{35.4} \\
        & \multicolumn{1}{l|}{\textbf{5.0e-5}} & \textbf{47.5} & 34.2 \\
        & \multicolumn{1}{l|}{2.0e-5} & 45.5 & 32.7 \\
        \midrule
        \multirow{3}{*}{\makecell[l]{UL loss weight $\alpha$}} 
        & \multicolumn{1}{l|}{5.0e-4} & 30.2 & 24.8 \\
        & \multicolumn{1}{l|}{3.0e-4} & 26.7 & 15.0 \\
        & \multicolumn{1}{l|}{\textbf{1.0e-4}} & \textbf{35.4} & \textbf{26.7} \\
        \midrule
        \multirow{3}{*}{\makecell[l]{PPL sampling interval}}
        & \multicolumn{1}{l|}{2.0-2.5} & 25.1 & 22.6 \\
        & \multicolumn{1}{l|}{\textbf{2.5-3.0}} & \textbf{27.4} & \textbf{23.6} \\
        & \multicolumn{1}{l|}{2.5-3.5} & 24.9 & 23.5 \\
        \midrule
        \multirow{3}{*}{\makecell[l]{$\sigma$ of Gaussian-guided\\ PPL sampling}}
        & \multicolumn{1}{l|}{0.5} & 33.5 & 24.5 \\
        & \multicolumn{1}{l|}{\textbf{0.25}} & \textbf{35.0} & \textbf{26.5} \\
        & \multicolumn{1}{l|}{0.1} & 31.8 & 26.4 \\
        \bottomrule
    \end{tabular}
    }
    \caption{Pass@1 scores of different configurations on AIME24 and AIME25 benchmarks.}
    \label{tab:ablation_study}
    \vspace{-0.7em}
\end{table}

\subsection{Ablation Study}
\label{sec:Ablation_Study}

\paragraph{$Q_1$: Whether each setup of OXA has been empirically verified? $A_1$: Yes.} Table \ref{tab:ablation_study} presents the results of ablating the hyperparameters used in OXA. By default, the experiments are conducted on the 1.5B model. The final configurations we use in the main experiment are bolded. Gaussian-guided PPL sampling outperforms the interval sampling, validating the benefit of mixing a small proportion of low-PPL data for better optimization.

\paragraph{$Q_2$: Can OXA models generalize to out-of-distribution reasoning tasks? $A_2$: Yes.} We further evaluate OXA models on the PhD-level problems via GPQA diamond \cite{DBLP:journals/corr/abs-2311-12022} and MMLU-STEM \cite{DBLP:conf/iclr/HendrycksBBZMSS21}. As illustrated in Figure \ref{fig:mix_fig2} (d), OXA models consistently outperform SFT baselines, demonstrating that our method effectively enhances the model's fundamental complex reasoning capabilities.

\begin{figure}
\centering
\includegraphics[width=0.45\textwidth]{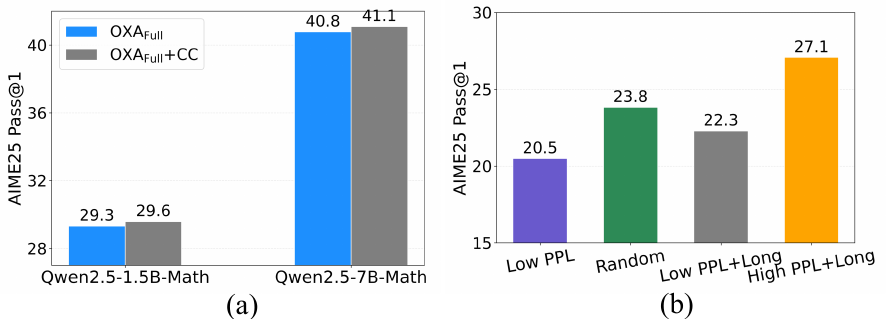}
\caption{(a) Results of OXA$_{\text{Full}}$ with Clip-Cov (OXA$_{\text{Full}}$+CC) on AIME25. (b) Comparison across different reasoning data selection strategies on AIME25. ``Low PPL'', ``Random'', and ``High PPL + Long'' correspond to SFT$_{\text{LP}}$, SFT, and OXA$_{\text{MLE}}$, respectively.}
\label{fig:mix_fig3}
\vspace{-0.9em}
\end{figure}

\paragraph{$Q_3$: Is OXA orthogonal to other methods? $A_3$: Yes.} We combine OXA with Clip-Cov \cite{DBLP:journals/corr/abs-2505-22617}, which is an RLVR-enhanced method that controls entropy by restricting the update of high-covariance tokens to encourage exploration. Figure \ref{fig:mix_fig3} (a) shows that OXA equipped with Clip-Cov achieves superior performance.

\paragraph{$Q_4$: Does OXA only benefit from selecting long data? $A_4$: No.} Figure \ref{fig:mix_fig3} (b) presents a comparison where the pre-trained LLM is fine-tuned on long data with low PPL. While this configuration yields marginal improvements over SFT$_{\text{LP}}$, it still lags significantly behind OXA$_{\text{MLE}}$, validating that the effectiveness of OXA stems from the integration of both low-confidence and long data.

\section{Conclusion}
We propose OXA to establish exploration-engaged initializations for the RLVR of mathematical LLMs. By leveraging MLE on low-confidence teacher data and unlikelihood training on high-confidence incorrect self-generated samples, OXA effectively boosts reasoning performance, expands the exploration space, and maintains high policy entropy. While validated on mathematics, OXA holds promise for other complex domains like code generation, which we defer to future work.

\section*{Limitations}
Despite the performance gains, our work has two primary limitations. First, compared to vanilla SFT, OXA$_{\text{Full}}$ incurs additional computational overhead due to the requirement of self-distillation reasoning trajectories. This sampling process is more resource-intensive than standard SFT. We further analyze the computation overhead of OXA in Appendix \ref{sec:Compute Overhead Discussion}. Second, due to limited computational resources, our empirical validation was restricted to models ranging from 1.5B to 7B parameters. However, we hypothesize that the benefits of OXA will be even more pronounced in larger-scale models. This is because larger models typically possess a greater capacity to internalize the reasoning paths with high PPL that smaller models might struggle to capture. We leave the exploration of OXA on larger model scales for future research.

\section*{Acknowledgments}
This work was supported in part by the NSFC (Nos. 62276056 and U24A20334). The authors would like to thank Yuhan Hou and Pengcheng Huang for their advice.

\bibliography{custom}

\appendix

\section{Appendix}

\subsection{Theoretical Analysis of Unlikelihood Loss}
\label{app:Theoretical Analysis of Unlikelihood Loss}

In this section, we analyze the gradient dynamics of the Unlikelihood Loss ($\mathcal{L}_{\text{UL}}$) compared to the standard Cross-Entropy Loss ($\mathcal{L}_{\text{CE}}$) to justify the necessity of a small scaling factor $\alpha$.

Let $z \in \mathbb{R}^{V}$ denote the logit vector output by the model at a specific time step $t$, where $V$ is the vocabulary size. Let $p_k = \text{softmax}(z)_k$ represent the predicted probability for token $k$. For a given input context $s_{<t}$, let $x_t$ be the specific token index targeted by the loss function (the ground truth token for $\mathcal{L}_{\text{CE}}$ or the negative token to be penalized for $\mathcal{L}_{\text{UL}}$).

\subsubsection{Gradient of Cross-Entropy Loss}
The gradient of $\mathcal{L}_{\text{CE}}$ with respect to any logit $z_j$ is bounded. Specifically:
\begin{align}
\frac{\partial \mathcal{L}_{\text{CE}}}{\partial z_j} = p_j - \mathbbm{1}(j = x_t),
\end{align}
where $\mathbbm{1}(\cdot)$ is the indicator function. Since $p_j \in (0, 1)$, the gradient magnitude is strictly bounded such that $\left| \frac{\partial \mathcal{L}_{\text{CE}}}{\partial z_j} \right| < 1$. This ensures stable updates regardless of the model's current confidence.

\subsubsection{Gradient of Unlikelihood Loss}
The unlikelihood objective aims to minimize the probability of a negative token $x_t$. The loss is defined as $\mathcal{L}_{\text{UL}} = - \log(1 - p_{x_t})$. Using the chain rule, we derive the gradient with respect to the logits.

For the target negative token logit ($z_{x_t}$):
\begin{align}
\frac{\partial \mathcal{L}_{\text{UL}}}{\partial z_{x_t}} &= \frac{\partial \mathcal{L}_{\text{UL}}}{\partial p_{x_t}} \cdot \frac{\partial p_{x_t}}{\partial z_{x_t}} \nonumber \\
&= \frac{1}{1 - p_{x_t}} \cdot p_{x_t}(1 - p_{x_t}) = p_{x_t}.
\end{align}
This term is bounded within $[0, 1]$. However, the instability arises from the gradients with respect to \textit{other} tokens $z_j$ (where $j \neq x_t$). The derivative of the softmax function for off-target indices is $\frac{\partial p_{x_t}}{\partial z_j} = -p_{x_t} p_j$. Thus:
\begin{align}
\frac{\partial \mathcal{L}_{\text{UL}}}{\partial z_j} &= \frac{\partial \mathcal{L}_{\text{UL}}}{\partial p_{x_t}} \cdot \frac{\partial p_{x_t}}{\partial z_j} \nonumber \\
&= \frac{1}{1 - p_{x_t}} \cdot (-p_{x_t} p_j) \nonumber \\
&= - p_j \cdot \underbrace{\left( \frac{p_{x_t}}{1 - p_{x_t}} \right)}_{\text{Odds Ratio Term}}.
\label{eq:ul_gradient_off_target}
\end{align}

\subsubsection{Instability and Weight Scaling}
Equation \ref{eq:ul_gradient_off_target} reveals a critical instability mechanism. The gradient for all non-target logits is scaled by the odds ratio of the negative token, $\frac{p_{x_t}}{1 - p_{x_t}}$.

\paragraph{High-Confidence Errors.} Consider a scenario where the model assigns a high probability to a hallucinated or incorrect token $x_t$ (e.g., $p_{x_t} = 0.99$). In this case, the gradient scaling factor becomes:
\begin{equation}
    \frac{0.99}{1 - 0.99} = 99.
\end{equation}
Consequently, the gradient applied to all other logits $z_j$ is amplified by a factor of roughly 100 compared to standard training dynamics. As $p_{x_t} \to 1$, this term approaches infinity.

If the scaling factor $\alpha$ in the final objective $\mathcal{L} = \mathcal{L}_{\text{CE}} + \alpha \cdot \mathcal{L}_{\text{UL}}$ is set to 1.0, these exploded gradients from high-confidence errors effectively overwrite the semantic knowledge learned via $\mathcal{L}_{\text{CE}}$, leading to catastrophic forgetting or model divergence. By setting $\alpha$ to a small value (e.g., $10^{-4}$), we counteract the explosion of the odds ratio term, ensuring that the unlikelihood updates remain comparable in magnitude to the maximum likelihood updates, thus stabilizing the training process.

\begin{figure*}
\centering
\includegraphics[width=1.0\textwidth]{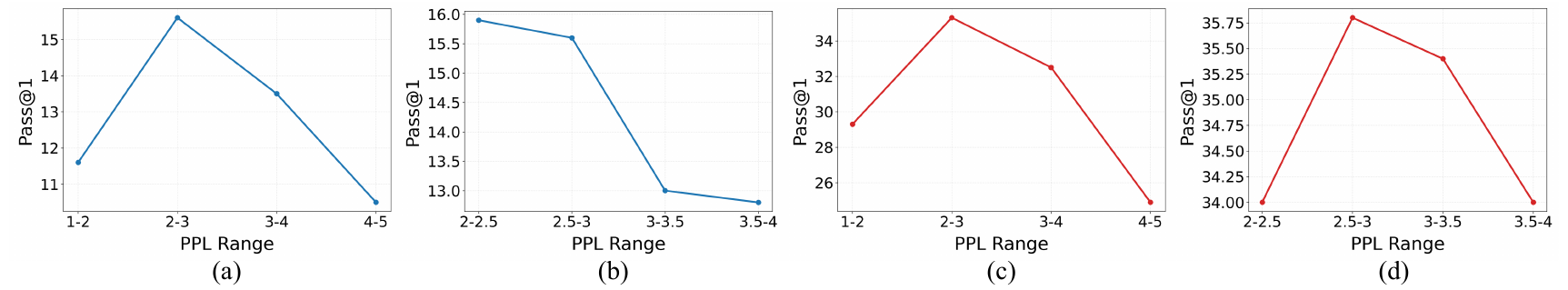}
\caption{Impact of data perplexity ranges on reasoning performance. (a)-(b) Performance on AIME24 of 1.5B models fine-tuned on various PPL intervals. (c)-(d) Corresponding results for 7B models across different PPL ranges.}
\label{fig:mix_fig4}
\end{figure*}

\subsection{Preliminary Experiments}
\label{app:Preliminary Experiments}

As illustrated in Figure \ref{fig:mix_fig4}, the performance of models fine-tuned on data subsets from different PPL ranges exhibits significant variance. Specifically, training on data with excessively low PPL (e.g., $1.0\text{--}2.0$) yields suboptimal reasoning capabilities, as these samples likely represent patterns the model has already mastered, offering limited signal for further improvement. In contrast, data with slightly higher PPL (e.g., $2.0\text{--}3.0$) achieves the best performance, as it trains the model to internalize previously uncaptured reasoning trajectories.

However, we also observe that if the PPL is too high, the complexity of the reasoning paths may exceed the model's current capacity, leading to optimization difficulties. To strike an optimal balance between learning and exploration potential, we propose Gaussian-guided PPL sampling. This approach allows us to concentrate training on the most effective PPL regions while maintaining a smooth distribution. Based on these preliminary findings, we set the Gaussian mean $\mu$ to $2.5$ for 1.5B models and $3.0$ for 7B models in our main experiments.

\subsection{Detailed Experimental Setup}
\label{app:Detailed Experimental Setup}

\paragraph{Model configurations.} To enable long-sequence modeling for mathematical reasoning, we adjust the rotary positional embedding (RoPE) parameters for the Qwen2.5-1.5B-Math and Qwen2.5-7B-Math models. Specifically, we increase rope theta from $10,000$ to $1,000,000$ and extend max position embeddings from $4,096$ to $40,000$. For \texttt{LLaMA3.2-3B} and \texttt{Qwen3-1.7B}, no modifications are required as their native context window of $32,768$ is sufficient for our experiments. Additionally, we remove the system prompt component from the tokenizer templates across all models to ensure a consistent and simplified input format.

\paragraph{Data preparation.} The first objective of OXA involves sampling from teacher-distilled SFT data. For \texttt{Qwen2.5-1.5B-Math}, the sampling hyperparameters are set as follows: \texttt{MIN PPL} = $1.0$, \texttt{MAX PPL} = $5.0$, \texttt{BIN WIDTH} = $0.05$, \texttt{TARGET STD} = $0.25$, and \texttt{TOTAL SAMPLES} = $50,000$. We maintain these settings for other models while adjusting the \texttt{TARGET CENTER}: it is set to $2.5$ for \texttt{Qwen2.5-1.5B-Math} and \texttt{LLaMA3.2-3B}, and $3.0$ for \texttt{Qwen2.5-7B-Math} and \texttt{Qwen3-1.7B}. To elicit structured chain-of-thought reasoning, we append the following instruction to each SFT problem: ``\textbackslash nLet's reason step by step. Enclose the reasoning process within <think>...</think>, then summarize it and present the final answer within \textbackslash boxed{} — for example: <think>reasoning process here</think> \textbackslash boxed{answer here}.'' For RLVR training, we use \texttt{DeepSeek-Distill-Qwen2.5-7B} to perform 8-sample generation per query. We select $10,000$ trajectories with pass rates between $0.2$ and $0.8$, effectively filtering out tasks that are either trivial or excessively difficult.

\paragraph{Training details.} For SFT, the Qwen2.5-1.5B-Math model is trained with a cutoff len of $32,768$, a learning rate of $2.5 \times 10^{-4}$, and $6.0$ epochs. We use a warmup ratio of $0.03$, weight decay of $0.1$, and Adam optimizer with $\beta_1=0.9, \beta_2=0.95$, using a global batch size of $128$. For other models, we adjust only the learning rates: $5.0 \times 10^{-5}$ for Qwen2.5-7B-Math, and $2.0 \times 10^{-4}$ for both \texttt{LLaMA3.2-3B} and \texttt{Qwen3-1.7B}. Reinforcement learning parameters are kept uniform: train batch size is $64$, max response length is $16,384$, actor learning rate is $2.0 \times 10^{-6}$, KL coefficient is $0.001$, with $8$ rollouts per query at a temperature of $0.85$. In our data scaling experiments with \texttt{Qwen2.5-7B-Math}, we increase the global batch size to $384$ and the learning rate to $1.5 \times 10^{-4}$ to ensure the total number of optimization updates remains consistent with our baseline experiments. We use \texttt{LLaMA-Factory} \cite{DBLP:journals/corr/abs-2403-13372} and \texttt{Verl} \cite{DBLP:conf/eurosys/ShengZYWZZPL025} for fine-tuning and reinforcement learning, respectively.

\subsection{Compute Overhead Discussion}
\label{sec:Compute Overhead Discussion}

Compared to conventional SFT, the additional computational overhead of OXA primarily stems from two components: PPL estimation for data selection and model decoding during self-distillation.

\paragraph{Efficiency of PPL estimation.} While OXA introduces a PPL calculation step, this process is highly efficient in practice. Since PPL estimation is performed through a single forward pass, the negative log-likelihoods of all tokens in a sequence are computed simultaneously in parallel. This allows the selection process to scale efficiently with sequence length, avoiding the sequential bottlenecks typical of autoregressive generation.

\paragraph{Self-distillation and mitigation.} The primary source of extra compute relative to vanilla SFT is the generation of high-confidence error trajectories via self-distillation. However, the cost of this phase is significantly mitigated by modern inference-time optimizations. In our pipeline, we leverage high-throughput inference frameworks such as vLLM \cite{DBLP:conf/sosp/KwonLZ0ZY0ZS23} and SGLang \cite{DBLP:conf/nips/ZhengYXS0YCKSGB24}, coupled with model quantization techniques. These advancements ensure that the generation of self-distilled data is both rapid and cost-effective, making OXA a practical choice for large-scale training.

\subsection{Comparison OXA Fine-Tuning with Direct Preference Optimization}
While both OXA and direct preference optimization (DPO) \cite{DBLP:conf/nips/RafailovSMMEF23} promote desirable samples and suppress undesirable ones, they differ fundamentally in two key aspects:

\paragraph{Data structure and coupling.} DPO is inherently a pairwise framework, requiring each training instance to consist of a triplet: a single query associated with both a ``chosen'' (desirable) and a ``rejected'' (undesirable) response. This constraint limits the utilization of unpaired data. In contrast, the data requirements for OXA are entirely decoupled. OXA permits queries to be associated with only a desirable or only an undesirable response, significantly lowering the barrier for data collection. Furthermore, this decoupling allows for flexible control over the mixing ratio of desirable and undesirable samples within each training batch, a hyperparameter that can be tuned to balance exploration and exploitation.

\paragraph{Optimization mechanism.} The training dynamics of the two methods are distinct. DPO employs a contrastive loss, which primarily focuses on maximizing the relative log-probability gap between the chosen and rejected responses. Conversely, OXA treats promotion and suppression as two independent objectives: the NLL loss focuses on internalizing correct reasoning patterns, while the unlikelihood loss directly minimizes the probability of incorrect paths. There is no intrinsic mathematical linkage between these two objectives in OXA, allowing the model to learn from each type of signal independently without the need for a direct comparison between two specific trajectories for every query.

\end{document}